# Image-Based Classification of Olive Species Specific to Türkiye with Deep Neural Networks

İrfan Atabaş[*1] Hatice Karataş[2]

[1,2]* Department of Computer Engineering, Faculty of Engineering and Natural Sciences, Kırıkkale University, KIRIKKALE

**Abstract:** In this study, image processing and deep learning methodologies were employed to automatically classify local olive species cultivated in Türkiye. A stereo camera was utilized to capture images of five distinct olive species, which were then preprocessed to ensure their suitability for analysis. Convolutional Neural Network (CNN) architectures, specifically MobileNetV2 and EfficientNetB0, were employed for image classification. These models were optimized through a transfer learning approach. The training and testing results indicated that the EfficientNetB0 model exhibited the optimal performance, with an accuracy of 94.5%. The findings demonstrate that deep learning-based systems offer an effective solution for classifying olive species with high accuracy. The developed method has significant potential for application in areas such as automatic identification and quality control of agricultural products.

## 1. Introduction

Olive is an agricultural product of considerable economic value both for table consumption and oil production [1]. Türkiye occupies a significant position in world olive production owing to its geographical location and climatic characteristics[1, 2]. Numerous local olive varieties with distinct morphological and physical features are cultivated across the Aegean, Mediterranean, Marmara, and Southeastern Anatolia regions. While this diversity reflects the country's agricultural richness, it simultaneously gives rise to challenges in terms of quality control, product tracking, and variety-based marketing strategies[3].

The accurate classification of olive varieties is a process that directly benefits both producers and. However, conventional classification methods are generally based on expert opinion and are susceptible to human error. Furthermore, since characteristics such as color, shape, and size of the olive fruit can be influenced by environmental factors over time, the manual evaluation of such parameters may not yield objective results. This has created a need for a faster, more reliable, and reproducible classification methodology.

In recent years, significant advances have been achieved in such classification tasks through the widespread adoption of image processing, artificial intelligence, and particularly deep learning technologies in agriculture[4-11]. Through image processing techniques, attributes such as shape, color, and texture of an object can be digitally quantified, and classification models can be constructed from these features using machine learning algorithms. In particular, deep learning models such as convolutional neural networks (CNNs) have come to the forefront in this field by achieving high accuracy rates in image classification problems[12, 13].

In this study, the classification of olive varieties grown in different regions of Türkiye using image processing and deep learning techniques was targeted. To this end, images of various olive varieties were collected using a stereo camera system, features were extracted from these images, and suitable classification models were developed. Due to the substantial computational power required during model training, a high-performance graphics processing unit (GPU)-supported hardware infrastructure was utilized throughout the project.

The primary contributions of this study are as follows:

- An image dataset comprising local olive varieties indigenous to Türkiye was constructed.

- Image processing-based feature extraction was performed, and various deep learning-based classification models were trained on the resulting data.
- Model performance was evaluated using metrics such as accuracy, precision, recall, and F1 score.
- An original system proposal for the automatic identification of olive varieties is presented.

The outcomes of this study may serve as a precursor to automated diagnostic systems applicable in production facilities or agricultural applications in the future. Furthermore, the classification results are expected to make significant contributions to the monitoring of agricultural product quality, the preservation of product standards, and the automation of production processes.

**2. Material and Method**

In this study, automatic discrimination of olive varieties grown in different regions of Türkiye was targeted through image processing techniques and deep learning-based classification methods. Within this scope, images of olive varieties were first collected using a stereo camera, after which these images were subjected to various preprocessing steps. In the final stage, classification was performed using Convolutional Neural Network (CNN) architectures. The methods followed from dataset construction to model training are presented in detail in this section. The schematic diagram of the study is provided in Figure 1.

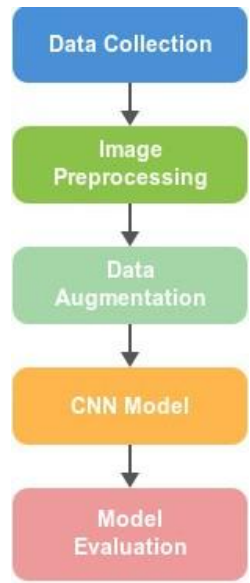

**Figure 1.** Schematic diagram of the proposed study

**2.1 Dataset Construction**

During the data collection phase, samples of five different local olive varieties grown in various agricultural regions of Türkiye were procured. Approximately 500 images per variety were captured at high resolution using a stereo camera system. Imaging was conducted under fixed lighting conditions and against a plain background to minimize the influence of environmental variables. The use of a stereo camera enabled the acquisition not only of two-dimensional images but also of depth information pertaining to individual olive fruits. This information facilitated the segmentation of images and enabled more precise delineation of object boundaries in subsequent stages. Representative sample images from the dataset are provided in Figure 2.

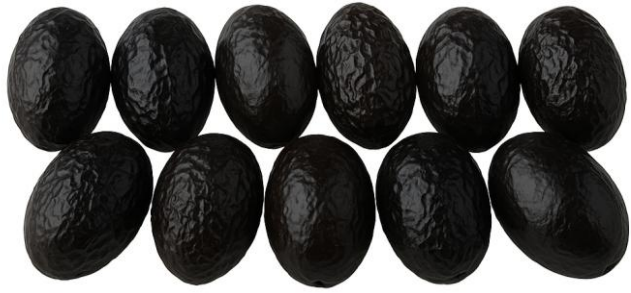

**Figure 2.** Representative sample images randomly selected from the dataset

**2.2 Image Preprocessing**

The collected images were subjected to various preprocessing steps to render them suitable for the classification task. First, all images were rescaled to a resolution of 224×224 pixels to conform to the input dimensions of the deep learning models. Subsequently, Gaussian filtering was applied to eliminate noise present in the images, thereby yielding sharper contours. A segmentation procedure was performed to reduce background influence and isolate the region occupied solely by the olive fruit. In this process, binary images were generated using Otsu's thresholding method, and only the object-specific regions were extracted through contour analysis. Data augmentation techniques were applied to enhance the model's ability to generalize. In this process, images were randomly rotated, flipped, and subjected to brightness variations to improve the model's robustness against diverse conditions.

**2.3 Convolutional Neural Network (CNN)**

CNNs are multilayer artificial neural network models that have revolutionized the field of image processing in particular. By extracting hierarchical features from input data, CNNs exhibit exceptional performance in learning complex patterns and structural relationships. The convolution operation enables the progressive abstraction of fundamental features such as edges, textures, and color transitions in images across successive layers, allowing the model to progress from simple visual cues to high-level semantic representations.

CNN architectures were first widely adopted in image classification and object recognition tasks. Particularly following AlexNet's remarkable success in the ImageNet competition in 2012, deep learning models became fundamental building blocks in the field of computer vision. Subsequently developed architectures such as VGG, ResNet, Inception, and EfficientNet achieved both deeper and more efficient structures, enabling their applicability to a wide range of problems[14, 15].

Today, CNN-based solutions are employed not only in image classification but also in various tasks including segmentation, object detection, face recognition, motion analysis, and scene interpretation[16]. Their effectiveness has been demonstrated in numerous application domains, including the analysis of radiological images for disease detection in medicine, product defect detection in industrial quality control processes, anomaly detection in security systems, traffic analysis in smart cities, and plant disease diagnosis in agriculture[17, 18].

In the agricultural domain, CNN-based models are employed for numerous tasks such as diagnosis of plant leaf diseases, fruit and vegetable classification, harvest prediction, and land classification; these models provide faster, more reproducible, and more objective results compared to conventional methods relying on human expertise[19-21].

In this context, CNN architectures constitute the foundation not only of theoretical successes but also of practical systems capable of generating solutions to real-world problems. For challenging classification problems such as the morphological similarity exhibited by olive varieties, the powerful feature extraction capabilities of CNNs offer considerable advantages. In this regard, CNN architectures have become not merely an option but a modern necessity in agricultural classification applications.

**2.4 Classification Model and Training**

The data obtained from the images were classified using deep learning methods. CNN architectures, which deliver high performance for image-based classification problems, were preferred. During model training, a transfer learning approach was adopted, and lightweight, fast-operating architectures previously trained on large datasets—specifically MobileNetV2 and EfficientNetB0—were integrated into the project. This approach reduced training time and enabled high accuracy rates to be achieved with limited data. The Adam optimizer was selected as the optimization algorithm during model training, and Categorical Cross-Entropy, which is well-suited to multi-class classification problems, was used as the loss function. The learning rate was set to 0.001, and training was conducted over 20 epochs. An early stopping strategy was implemented to prevent the model from overfitting.

The obtained dataset was partitioned into three subsets: training (80%), validation (10%), and testing (10%). Following the training process, the optimal weights based on validation performance were recorded. The final model was evaluated on the test dataset using performance metrics such as accuracy, recall, specificity, and F1 score.

## 3. Results

This section presents the experimental evaluations and the performance metrics obtained from the developed classification models. The models trained on images belonging to different olive varieties were tested on the previously partitioned test dataset. The obtained results were analyzed using widely adopted classification performance metrics including accuracy, precision, recall, and F1 score.

Two different deep learning architectures were comparatively employed in the study: MobileNetV2 and EfficientNetB0. Using a transfer learning approach, these models were customized to achieve high accuracy with limited data. The performance evaluation results of the models are presented in Table 1.

**Table 1.** Classification performance of MobileNetV2 and EfficientNetB0 models

| Model | Accuracy (%) | Precision | Recall | F1 Score |
| --- | --- | --- | --- | --- |
| MobileNetV2 | 92.8 | 0.91 | 0.93 | 0.92 |
| EfficientNetB0 | 94.5 | 0.94 | 0.95 | 0.94 |

The results indicate that both models were able to classify olive varieties with high performance. However, the EfficientNetB0 architecture demonstrated superior performance across all metrics compared to MobileNetV2. This difference may be attributed to the more detailed feature extraction capability and the broader learning capacity of the model.

The classification accuracies of the models were monitored throughout the training process, and the accuracy/epoch curves are presented in Figure 3. In both models, accuracy values increased steadily, no signs of overfitting were observed, and the early stopping mechanism was found to operate effectively.

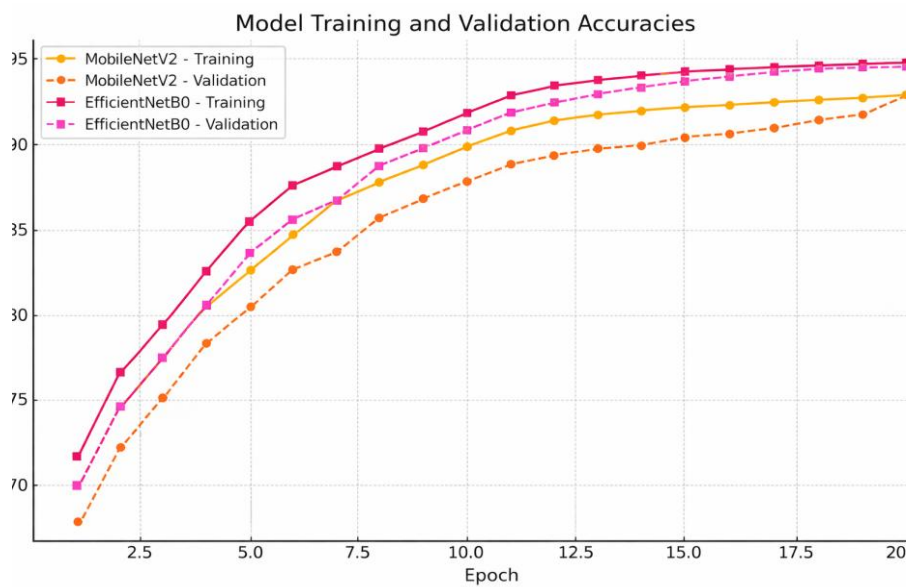

**Figure 3.** Training and validation accuracy curves (MobileNetV2 and EfficientNetB0)

Additionally, a confusion matrix was generated to analyze classification performance in greater detail. The matrix constructed for the EfficientNetB0 model revealed that high accuracy was achieved across all classes and that discrimination between visually similar varieties was successfully accomplished. Although minimal misclassifications were observed between certain classes, the overall classification performance remained above 94%.

In conclusion, the developed deep learning-based system enabled the classification of olive varieties grown in Türkiye with high accuracy. These results demonstrate that the proposed approach can offer a reliable and practical solution for the agricultural sector.

**4. Discussion and Conclusion**

In this study, the classification of local olive varieties grown in Türkiye using image processing and deep learning methods was targeted, and the results obtained were found to be highly successful when compared to similar studies in the literature. The use of deep learning architectures provided higher accuracy and generalization capability compared to conventional machine learning methods, once again demonstrating the effectiveness of CNN-based models in agricultural product classification.

The preference for stereo camera technology in the construction of the dataset was assessed as one of the significant advantages of the study. Thanks to this technology, not only two-dimensional images of olive fruits but also three-dimensional structural characteristics were incorporated into the dataset, and this positively impacted segmentation and classification performance. However, it should also be considered that the processing of stereo images requires a more complex procedure compared to classical single-view images.

The preprocessing and data augmentation steps were observed to play a critical role in model performance. In particular, the augmentation of images captured from different angles and under various lighting conditions helped the model achieve generalized learning without overfitting. The variety of data augmentation techniques employed at this stage made it possible for the model to become more adaptable to different scenarios that may be encountered in field conditions.

In the comparison conducted among lightweight, fast-operating CNN architectures such as MobileNetV2 and EfficientNetB0, the EfficientNetB0 model was observed to be superior in all performance metrics. This result can be attributed to the optimized layer structure and parametric efficiency of the EfficientNet architecture. The achievement of high accuracy rates despite a short training duration demonstrates the effectiveness of the transfer learning method and confirms that it offers an important solution for studies with limited data such as this.

Nevertheless, the observation of classification errors between certain varieties indicates that the discrimination of morphologically similar olive varieties by the model can still pose a challenge. It is thought that this issue can be overcome in future studies through larger sample sizes and more advanced architectures. Furthermore, the integration of the model into mobile or embedded systems to enhance its field applicability can also be considered an important research direction.

In future studies, the dataset may be expanded to further improve the generalization capability of the model, and additional olive varieties may be incorporated into the classification process. Moreover, it is possible to develop more robust models by increasing feature diversity through different imaging techniques such as hyperspectral imaging. In addition, integration with mobile devices or embedded systems can be achieved to enable the use of the developed system under field conditions or in production facilities.

In conclusion, this study has demonstrated that image processing and deep learning techniques can be effectively employed in the classification of olive varieties, while also once again highlighting the power and applicability of artificial intelligence-based solutions in agricultural technologies.

**Acknowledgment**

This study was carried out within the scope of project number 2023/152 supported by the Kırıkkale University Scientific Research Projects Coordination Unit (BAP). The author would like to express sincere gratitude to the Kırıkkale University BAP Commission for their support.